\documentclass[10pt,twocolumn,letterpaper]{article}

\usepackage[pagenumbers]{cvpr} 

\definecolor{cvprblue}{rgb}{0.21,0.49,0.74}
\usepackage[pagebackref,breaklinks,colorlinks,allcolors=cvprblue]{hyperref}
\usepackage{hyperref}
\usepackage{graphicx}
\graphicspath{{./images/}}
\usepackage{multirow}

\usepackage{algorithm}
\usepackage{alphalph}
\usepackage{algorithmic}
\usepackage{bbding}
\usepackage{makecell}
\usepackage[accsupp]{axessibility}

\usepackage{orcidlink}

\usepackage{enumitem}

\usepackage{subcaption}

\usepackage[utf8]{inputenc} 
\usepackage[T1]{fontenc}    
\usepackage{url}            
\usepackage{booktabs}       
\usepackage{amsfonts}       
\usepackage{nicefrac}       
\usepackage{microtype}      
\usepackage{amssymb}
\usepackage{pifont}

\usepackage[table,dvipsnames]{xcolor}
\definecolor{LightCyan}{rgb}{0.91,0.91,0.98}
\definecolor{LightYellow}{rgb}{1.0, 1.0, 0.88}
\definecolor{magicmint}{rgb}{0.67, 0.94, 0.82}
\definecolor{lightmauve}{rgb}{0.86, 0.82, 1.0}
\definecolor{grannysmithapple}{rgb}{0.66, 0.89, 0.63}
\definecolor{isabelline}{rgb}{0.95,0.93,0.91}

\def\paperID{16367} 
\def\confName{CVPR}
\def\confYear{2026}

\title{OAMVOS:2nd Report for 5th PVUW MOSE Track}
\author{
Deshui Miao$^{1}$ \quad
Xingsen Huang$^{1}$ \quad
Yameng Gu$^{1}$ \\
Xiaogang Yu$^{2}$ \quad
Xin Li$^{1*}$ \quad
Ming\mbox{-}Hsuan Yang$^{3}$ \\
$^{1}$Pengcheng Laboratory \\
$^{2}$Guangzhou Hengyan Technology \\
$^{3}$University of California at Merced \\
}

\begin{document}
\maketitle
\begin{abstract}
SAM-based dense trackers provide strong short-term mask propagation but remain fragile under long occlusion, fast motion, viewpoint change, and distractors. The problem is especially severe for small objects, where a few incorrect memory updates can dominate later predictions. This report presents an occlusion- and reappearance-aware extension of DAM4SAM that improves memory control rather than changing the backbone. The method augments the original SAM3 tracker with four ingredients: a reliability-aware tracking state machine, branch-based recovery, delayed DRM promotion, and a selective policy for native SAM3 memory selection. During stable tracking, the model follows the original single-path propagation process. Once confidence drops, the tracker enters an ambiguous or recovery mode, maintains a small set of candidate branches, and commits memory only after a branch is reconfirmed. For small-object disappearance and reappearance, native memory selection is temporarily bypassed so older anchors remain accessible. In addition, the first conditioning frame is explicitly preserved, and the conditioning-memory budget is moderately enlarged to improve long-gap recovery. The resulting design keeps DAM4SAM efficient in easy cases while improving robustness in sequences dominated by occlusion and reappearance.
\end{abstract}

\section{Introduction}
Tracking-by-segmentation systems based on foundation models have improved prompt flexibility and mask quality, but robustness under long-range temporal variation is still limited\cite{tokmakov2023breaking,liu2024grounding,oquab2023dinov2, ding2025mosev2}. In practice, three failure modes dominate complex video object segmentation: fast motion~\cite{cheng2024putting}, distractor-rich scenes~\cite{huang2019got, ding2025mosev2}, and reappearance after long occlusion~\cite{fan2019lasot} or strong viewpoint change. These cases are even harder for small objects because mask geometry is unstable, the object pointer becomes noisy, and a small localization error can erase the target from memory~\cite{yang2021associating, ding2025mosev2}.
CVPR 2026 5th PVUW challenge has three tracks: Complex VOS on
MOSEv2~\cite{ding2025mosev2}, targeting realistic, cluttered scenes with
small, occluded, reappearing, and camouflaged objects
under adverse conditions; VOS on MOSE~\cite{ding2023mose}, focusing
on challenging, long, and diverse videos; and RVOS on
MeViS~\cite{ding2023mevis, MeViSv2}, assessing referring video object segmentation with text or audio.

DAM4SAM~\cite{DAM4SAM} alleviates part of this problem by promoting selected frames into Dynamic Anchor Memory (DRM), allowing informative historical views to act as conditioning anchors. However, the original pipeline is still mostly single-path: predictions are generated, committed, and later reused as memory in a tightly coupled way. Under severe ambiguity, one incorrect commit can contaminate subsequent memory and trigger compounding drift~\cite{kristan2022tenth, kristan2023first,kristan2024second}. We also observed that native SAM3~\cite{carion2025sam} memory selection, while useful in stable tracking, can harm small-object reappearance because it overemphasizes recent low-quality memories from occlusion periods and suppresses older but semantically correct anchors.

The central idea of this project is that robust long-term tracking depends more on memory governance than on a new segmentation backbone~\cite{zhou2024rmem,ravi2024sam}. We therefore keep the original SAM3 tracker and DAM4SAM memory encoder, but add a control layer that explicitly distinguishes stable tracking from uncertain tracking. Stable frames may update reference statistics and promote new DRM entries. Uncertain frames are treated as provisional: they are explored through a small branch pool and are not allowed to directly dominate long-term memory until consistency is restored.

The resulting design combines appearance-aware anchors, motion and geometry consistency, branch-based recovery, delayed DRM promotion, conditional use of native SAM3 memory selection, and explicit preservation of the first conditioning frame during attention. The rest of this report formalizes these components.

\section{Methods}
\begin{figure*}
    \centering
    \resizebox{0.9\linewidth}{!}{
        \includegraphics{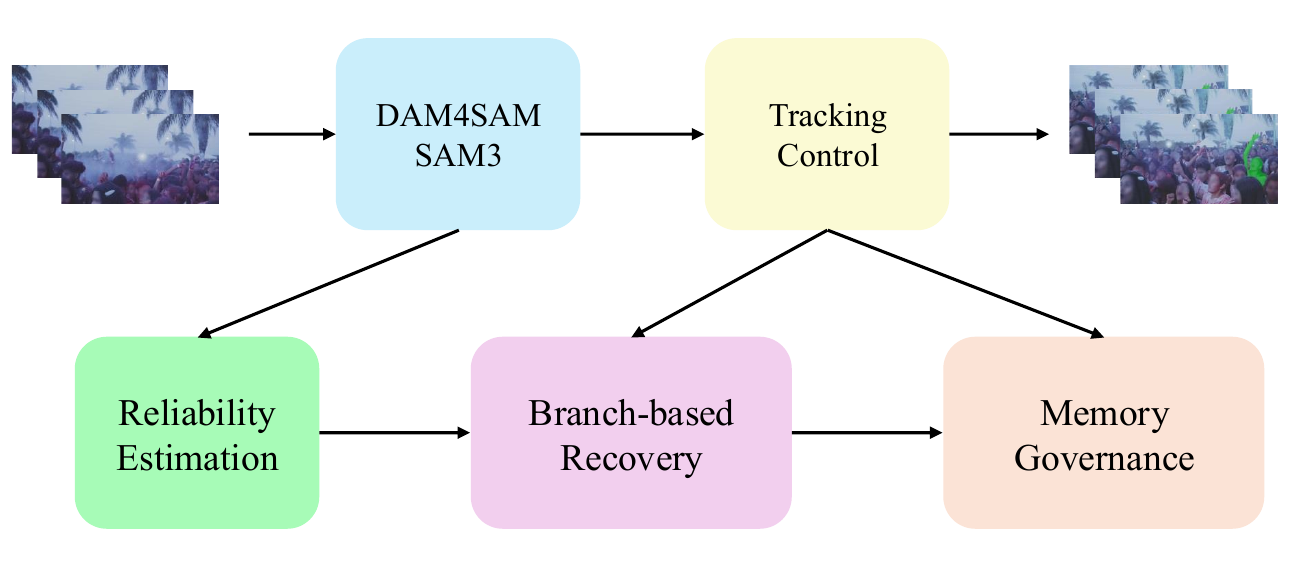}
    }
    \caption{\textbf{Pipeline of our methods.}}
    \label{fig:mevis_audio}
     \vspace{-4mm}
\end{figure*}
\subsection{Overview}
The method is built on top of the SAM3-based DAM4SAM tracker. Let $I_t$ denote frame $t$, $m_t$ the predicted mask, and $p_t \in \mathbb{R}^d$ the corresponding object pointer. After initialization, each frame is processed in one of three modes:
\begin{equation}
z_t \in \{\texttt{stable},\texttt{ambiguous},\texttt{recovery}\}.
\end{equation}
In the stable mode, the tracker follows the original DAM4SAM path and performs speculative prediction on the current frame. Reliable predictions are committed as non-conditioning outputs and may later be promoted into DRM. Unreliable predictions trigger an uncertain regime, where several candidate branches are propagated independently, and only a reconfirmed branch is allowed to return to the main path. This separation between candidate generation and memory commitment is the key modification.

\subsection{Reliability Estimation in Stable Tracking}
For every tracked frame, the base predictor returns a primary mask, optional alternative masks, objectness logits, and predicted IoU values. We convert these outputs into four interpretable scores.

\paragraph{Appearance score.}
We maintain an anchor bank $\mathcal{A}=\{a_i\}_{i=1}^{N_A}$ formed from the initialization frame and later verified high-confidence frames. After $\ell_2$ normalization, the appearance score is
\begin{equation}
s_t^{\text{app}}=\max_{a_i \in \mathcal{A}} \frac{\langle p_t, a_i \rangle + 1}{2},
\end{equation}
which maps cosine similarity into $[0,1]$.

\paragraph{Motion score.}
Let $c_t$ be the center of the predicted mask. Using the last two stable centers, we estimate a constant-velocity prediction
\begin{equation}
\hat{c}_t = c_{t-1} + (c_{t-1}-c_{t-2}),
\end{equation}
and define
\begin{equation}
s_t^{\text{mot}} = \exp\!\left(-\frac{\|c_t-\hat{c}_t\|_2}{\tau_m}\right).
\end{equation}
This term penalizes implausible jumps induced by fast motion or distractor switches.

\paragraph{Geometry score.}
Let $A_t = |m_t|$ denote mask area and $\rho_t$ its aspect ratio. Using medians $\bar{A}$ and $\bar{\rho}$ from stable history, we compute
\begin{equation}
r_t^{A}=\min\!\left(\frac{A_t}{\bar{A}},\frac{\bar{A}}{A_t}\right), \quad
r_t^{\rho}=\min\!\left(\frac{\rho_t}{\bar{\rho}},\frac{\bar{\rho}}{\rho_t}\right),
\end{equation}
and combine them as
\begin{equation}
s_t^{\text{geo}} = \text{clip}\!\left(0.7\,r_t^{A}+0.3\,r_t^{\rho},0,1\right).
\end{equation}
For very small objects, the area term is relaxed to reduce sensitivity to minor mask fluctuations.

\paragraph{Candidate margin.}
If the predictor returns candidate IoU values $\{q_t^{(k)}\}_{k=1}^{K}$, we define
\begin{equation}
\Delta_t = q_t^{(1)} - q_t^{(2)},
\end{equation}
where $q_t^{(1)} \ge q_t^{(2)}$ are the two largest values.

These cues are combined with the top IoU $q_t=q_t^{(1)}$ to determine the tracking mode:
\begin{equation}
z_t=
\begin{cases}
\texttt{recovery}, & A_t=0 \ \text{or}\ q_t<\tau_{\text{rec}} \ \text{or}\ s_t^{\text{app}}<\tau_{\text{app,rec}},\\[0.3em]
\texttt{ambiguous}, & q_t<\tau_{\text{unc}} \ \text{or}\ s_t^{\text{app}}<\tau_{\text{app,unc}}\\
& \hspace{1.1em}\text{or}\ s_t^{\text{mot}}<\tau_{\text{mot}} \ \text{or}\ s_t^{\text{geo}}<\tau_{\text{geo}} \ \text{or}\ \Delta_t<\tau_{\Delta},\\[0.3em]
\texttt{stable}, & \text{otherwise}.
\end{cases}
\end{equation}
This stage is not deciding whether to store a frame in memory; it is deciding whether the current single-path state is trustworthy enough to be committed.

\subsection{Branch-Based Recovery}
Once a frame is marked uncertain, the tracker constructs a branch pool
\begin{equation}
\mathcal{B}_t = \{b_t^{(1)},\ldots,b_t^{(N_t)}\},
\end{equation}
initialized from the current predictor output. A branch may correspond to the primary mask, one of the strongest alternative masks, or an explicit absent-object hypothesis. Each branch owns an independent copy of the inference state, so hypotheses can evolve without polluting the main path. For non-primary masks, the candidate mask is re-injected as a prompt into the predictor to obtain a branch-specific memory state and object pointer.

Branch scores accumulate evidence over time rather than relying on a single frame. For branch $b$ at frame $t$, we define
\begin{equation}
\label{eq:branch_score}
\begin{aligned}
S_t(b)=\;&S_{t-1}(b)+\log(\max(q_t(b),\varepsilon))
+\frac{1}{2}\log(\max(o_t(b),\varepsilon)) \\
&+\lambda_a s_t^{\text{app}}(b)
+\lambda_m s_t^{\text{mot}}(b)
+\lambda_g s_t^{\text{geo}}(b) \\
&-\lambda_e\,\mathbb{1}[A_t(b)=0].
\end{aligned}
\end{equation}
where $o_t(b)$ is the objectness probability and $\varepsilon$ avoids numerical instability. In our implementation, appearance receives the largest auxiliary weight because identity preservation matters more than local geometry during recovery.

After expansion, only the best branch per root identity is kept, and the global pool is truncated to the top few hypotheses. Recovery ends when one branch wins consistently and satisfies a reconfirmation rule. For the generic case, we require several consecutive wins and strong appearance agreement. For reappearance after a missing streak of length at least $L_{\text{miss}}$, we use a relaxed rule
\begin{equation}
\begin{aligned}
\mathbf{1}_{\mathrm{reconfirm}}
=\;&\mathbf{1}[n_{\mathrm{win}}\ge 1]
\cdot \mathbf{1}[q_t \ge \tau_{\mathrm{rep,iou}}]
\cdot \mathbf{1}[s_t^{\mathrm{app}}\ge \tau_{\mathrm{rep,app}}] \\
&\cdot \mathbf{1}[\Delta_t \ge \tau_{\mathrm{rep},\Delta}]
\cdot \mathbf{1}[A_t>0].
\end{aligned}
\end{equation}
This relaxation is important for small objects, which often return with weak geometry in the first visible frame.

\subsection{Anchor Bank and Delayed DRM Promotion}
The method maintains two complementary long-term structures: an anchor bank and DRM. The anchor bank stores normalized object pointers from the initialization frame and later high-confidence stable frames. DRM, by contrast, is the stronger conditioning memory already supported by DAM4SAM and is used directly by the transformer encoder during future frame processing.

Not every stable prediction should be promoted into DRM. We therefore use delayed promotion. Let $g_t$ denote the frame gap to the most recent DRM insertion and let $r_t$ denote the current-to-reference size ratio. A stable frame becomes a DRM candidate if
\begin{equation}
\mathbb{1}[\text{cand}_{\text{drm}}] =
\mathbb{1}[q_t \ge \tau_{\text{drm}}]\,
\mathbb{1}[g_t \ge G_{\min}]\,
\mathbb{1}[r_t \in [r_{\min},r_{\max}]]
\end{equation}
for normal objects, with relaxed size constraints for small objects. For non-reappearance frames, promotion also requires a distractor signal extracted from alternative masks, indicating that the frame captures a useful disambiguating view. For reappearance, the IoU threshold is relaxed because the first newly visible frame is often highly informative.

Even when a frame qualifies as a candidate, promotion is not immediate. Let $u_t$ be the number of consecutive frames satisfying the promotion criterion. DRM insertion is performed only when
\begin{equation}
\mathbb{1}[\text{add to DRM}] = \mathbb{1}[u_t \ge N_{\text{drm}}],
\end{equation}
which prevents transient recovery noise from entering the strongest memory bank.

\begin{table*}[t]
\centering
\caption{Leaderboard results.}
\label{tab:leaderboard}
\begin{tabular}{c l c c c c c c c}
\toprule
Rank & Participant & J\&F & J & F & N-acc. & T-acc. & Score-1 & Final \\
\midrule
1 & HITsz\_Dragon & 56.91 & 54.71 & 59.12 & 62.09 & 44.11 & 62.39 & 58.55 \\
2 & \textbf{tobedone}      & 55.16 & 53.02 & 57.29 & 58.51 & 44.47 & 61.02 & 57.02 \\
3 & mmm           & 51.17 & 49.55 & 52.80 & 66.23 & 32.60 & 55.13 & 52.34 \\
4 & newsota       & 48.19 & 46.16 & 50.22 & 55.36 & 37.21 & 52.97 & 49.56 \\
5 & rozumrus      & 44.83 & 43.18 & 46.48 & 55.81 & 31.29 & 48.96 & 46.07 \\
\bottomrule
\end{tabular}
\end{table*}

\subsection{Conditional Use of Memory Selection}
Native SAM3 memory selection filters non-conditioning memories using a quality score derived from objectness and predicted IoU. This is useful when recent frames are informative and clean. However, for small-object disappearance and reappearance, recent memories are often dominated by occlusion or near-empty observations, so aggressive filtering can suppress the older anchors needed for recovery.

We therefore disable native memory selection only in the specific regime where it is harmful. Let $\sigma_t$ denote whether the object is considered small and let $m_t^{\text{miss}}$ denote the current missing streak. The bypass indicator is
\begin{equation}
\gamma_t=
\mathbb{1}[\sigma_t=1]\,
\mathbb{1}[m_t^{\text{miss}}>0 \ \lor\ z_t \neq \texttt{stable}].
\end{equation}
The actual selection mode is then
\begin{equation}
\texttt{use\_memory\_selection}_t = 1-\gamma_t.
\end{equation}
Hence native selection remains active during normal stable tracking and is bypassed only during small-object disappearance, ambiguity, or recovery.

\subsection{Conditioning-Memory Attention for Reappearance}
The base SAM3 tracker limits the number of conditioning frames participating in attention through a parameter $K_c=\texttt{max\_cond\_frames\_in\_attn}$. This is important for efficiency but can hurt long-gap reappearance when the first conditioning frame is discarded in favor of temporally closer DRM entries. We therefore preserve the first conditioning frame explicitly and modestly enlarge the attention budget.

Let $\mathcal{D}_t$ denote the set of available DRM frames before processing frame $t$, and let frame $0$ be the initialization frame. The conditioning set is constructed as
\begin{equation}
\mathcal{C}_t = \{0\} \cup \text{TopK}_{j \in \mathcal{D}_t \setminus \{0\}}\big(-|j-t|, K_c-1\big).
\end{equation}
This guarantees that the clean initial target definition is always eligible, while the remaining slots are filled with temporally relevant DRM frames. In our current configuration, $K_c$ is increased from $4$ to $6$.

The final pipeline is simple. The first frame initializes the tracker, reference geometry, and anchor bank. Each new frame stays on the stable single-path route until reliability degrades. Once uncertainty appears, the tracker enters ambiguous or recovery mode, evaluates a small set of hypotheses, and commits only a reconfirmed branch. Small-object reappearance temporarily disables native memory selection so that old anchors remain accessible. High-value frames are promoted into DRM through a delayed policy, and the first conditioning frame is always preserved in conditioning attention. The key idea is therefore not a new backbone, but a stricter policy for when predictions are trusted and how memory is used during failure and recovery.

\section{Experiments}

We evaluate the proposed method on this challenge that emphasizes long-term ambiguity, disappearance, and recovery. 
The proposed method achieves second place in this challenge.
Our implementation uses the SAM3-based DAM4SAM tracker with `max\_cond\_frames\_in\_attn = 6`, `keep\_first\_cond\_frame = True`, `BRANCH\_KEEP = 3`, delayed DRM promotion, and the small-object reappearance bypass for native SAM3 memory selection. The tracker enters `stable`, `ambiguous`, or `recovery` mode according to the reliability signals described in Section 2, and DRM updates are performed only after the delayed-promotion criterion is satisfied. 
{
    \small
    \bibliographystyle{ieeenat_fullname}
    \bibliography{main}
}


\end{document}